\documentclass{article}
\usepackage[final, nonatbib]{dlde_neurips_2021}
\usepackage{algorithm}
\usepackage{algpseudocode} 
\usepackage{graphicx}
\usepackage{amsmath}
\newcommand{\norm}[1]{\left\lVert#1\right\rVert}
\usepackage[colorlinks,urlcolor=blue,linkcolor=blue,citecolor=blue]{hyperref}
\usepackage{color,array}
\usepackage{multirow}
\usepackage{booktabs}
\usepackage{caption}
\usepackage{subcaption}
\usepackage{xcolor}         

\title{LiFe-net: Data-driven Modelling of Time-dependent
Temperatures and Charging Statistics Of Tesla’s
LiFePo4 EV Battery}
\author{Jeyhun Rustamov \\ Helmholtz-Zentrum Dresden Rossendorf\\
Bautzner Landstraße 400, 01328 Dresden, Germany \\ e-mail: j.rustamov@hzdr.de\\
\And 
Luisa Fennert \\ TU Chemnitz \\ Str. der Nationen 62, 09111 Chemnitz,
Germany \\ e-mail: luisa.fennert@s2021.tu-chemnitz.de
\And
Nico Hoffmann \\ Helmholtz-Zentrum Dresden Rossendorf \\ Bautzner
Landstraße 400, 01328 Dresden, Germany \\ e-mail: n.hoffmann@hzdr.de
}
\date{August 2022}

\begin{document}

\maketitle

\begin{abstract}
Modelling the temperature of Electric Vehicle (EV) batteries is a fundamental task of EV manufacturing.
Extreme temperatures in the battery
packs can affect their longevity
and power output. Although theoretical models exist for describing heat transfer in battery packs, they are
computationally expensive to simulate. Furthermore, it is difficult to acquire data measurements from within the battery cell. In this work, we propose a data-driven surrogate model (LiFe-net) that uses readily accessible driving diagnostics for battery temperature estimation to overcome these limitations. This model incorporates Neural Operators with a traditional numerical integration scheme to estimate the temperature evolution. Moreover, we propose two further variations of the baseline model: LiFe-net trained with a regulariser and LiFe-net trained with time stability loss. We compared these models in terms of generalization error on test data. The results showed that LiFe-net trained with time stability loss outperforms the other two models and can estimate the temperature evolution on unseen data with a relative error of 2.77 \% on average.  

\end{abstract}

\section{Introduction}

As a result of ongoing attempts to decrease the dependency on fossil fuels, the public interest in electric vehicles (EVs) has been steadily growing for the past decade. Consequently, the need for efficient batteries has also increased. In this work, we examine the Lithium Iron Phosphate (LiFePO$_{4}$) battery used in the Tesla Model 3. LiFePO$_{4}$ batteries are a category of Lithium-ion (Li-ion) batteries which are widely used due to their high energy density, small size, and wide operating temperature range \cite{Ismail_2013}, \cite{Ling2015}, \cite{Tran2022}. However, they are very sensitive to temperature variations \cite{Mathewson14}. At high temperatures, Li-ion batteries degrade rapidly, while the low temperatures cause a reduction in power and energy output \cite{Pals95}. Therefore, monitoring temperature  evolution is crucial for EV manufacturers.  Theoretically, heat transfer inside Li-ion batteries can be described by Equation \ref{eq:34} \cite{Ismail_2013},\cite{Rad2013}.
\begin{equation}
\label{eq:34}
\begin{aligned}
    mc_{cell}\dfrac{\mathrm{d}T_{cell}}{\mathrm{d}t} = I^2 R + T_{cell} \Delta S \dfrac{I}{nF} + Ah(T_{cell} - T_{amb}),
\end{aligned}
\end{equation}
The first part represents the heat generated by the battery cell's internal resistance. The second term denotes the entropy change during discharge. Finally, the last term is the heat transferred to the ambience by convection \cite{Ismail_2013}. Traditional numerical methods for solving such problems require a new simulation for every new instance of the ODE and hence are computationally expensive. Therefore, there is a potential for alternative data-driven models that are fast, flexible, and robust in predicting battery temperature. Motivated by this potential, we make following contributions in this work:
\begin{enumerate}
    \item We propose a data-driven approach for learning time and environmental condition-dependent neural operator that predict the temperature evolution of a LiFePO$_4$ EV battery.
    \item We introduce regularisation strategies to improve the robustness of the neural operator \cite{Kovachki2021}.
    \item The connection of neural operators with a surrogate model allows for in-situ prediction of charging statistics.
    \item Good generalisation is verified in terms of unseen driving data.
\end{enumerate}

\section{Methodology}
The experimental setup for acquiring the battery temperature data is shown in Fig.\ref{fig:sensorinst} in Appendix A. The temperature probe of an Inkbird IBS-TH1 was installed on the floor of the cabin of a Tesla Model 3 with a CATL LiFePo$_{4}$ battery under the driver’s seat. This setup also offers insights into the electrochemical system such as maximum charging power. We also collected different driving diagnostics from Tesla API, including the car's speed, power
consumption, outside temperature, and battery level.
\subsection{Neural Operator}
We model the baseline model of predicting battery temperature evolution in the form of Ordinary Differential Equations (ODEs) as shown in Equation \ref{eqn:ode} :
\begin{equation}
\begin{aligned}
	\dfrac{\mathrm{d}u(t,\textbf{E})}{\mathrm{d}t}= f(t, u(t,\textbf{E})) \approx NN_{\theta}(t, \textbf{E}), \qquad t \in [0, T]  \\
	u(t_{0}, \textbf{E}_{0} ) = u_{0} \qquad \qquad \qquad \qquad \qquad
\end{aligned}
\label{eqn:ode}
\end{equation}
where u(t, \textbf{E}) is our hidden solution of temperature, t is relative time (duration of each drive session starting from $t_{0}$ = 0 to final time $T$), \textbf{E} is the vector of environmental parameters, while $u_{0}$ and \textbf{E$_{0}$} are the initial temperature and initial environmental parameters, respectively.
We represent the unknown right-hand side $f(t, u(t,\textbf{E})$ with a neural network $NN_{\theta}$ (multilayer perceptron (MLP)) as shown in Equation \ref{eqn:ode}.
It takes relative time and environmental parameters (power, speed, battery level, outside temperature, current battery temperature) as inputs and outputs the estimation of the neural operator at discrete time steps. 
The neural network is then trained by minimizing the mean squared error (MSE) between the prediction of our network and the ground-truth values for the differential operator calculated by forward differences (FD) as shown in Equation \ref{eqn:mseloss}.
Thus, we learn the neural operator that describes the underlying physics of time-dependent temperature evolution in a supervised manner.
\begin{equation}
\begin{aligned}
	\mathcal{L}_{NO} = \sum_{i=0}^{N-1} \ \norm{ \dfrac{u(t_{i+1},\textbf{E$_{i+1}$})-u(t_{i},\textbf{E$_{i}$})}{t_{i+1}-t_{i}} - NN_{\theta}(t_{i}, \textbf{E}_{i}) }_2^2 \\[2ex]
\end{aligned}
\label{eqn:mseloss}
\end{equation}
The trained neural operator is then used in Euler Forward numerical scheme to iteratively estimate the complete temperature evolution during inference with the time-step $h$ as shown in Equation \ref{eqn:euler}:
\begin{equation} 
\begin{aligned}
    u(t_{i+1}, \textbf{E$_{i+1}$}) = u(t_{i}, \textbf{E$_{i}$}) + h* NN_{\theta}(t_{i}, \textbf{E}_{i}) \qquad for \quad i = 0, \dots , N-1
	\\[2ex]
\end{aligned}
\label{eqn:euler}
\end{equation}
\subsection{Regularisation}
To explore further improvements to the baseline model, we introduce a regularizer $\mathcal{L}_{smooth}$ minimizing the norm of the partial derivative of $NN_{\theta}$  with respect to its input parameters to improve the smoothness of the prediction (see Equation \ref{eqn:reg}).
\begin{equation}
\begin{aligned}
	\mathcal{L}_{smooth} = \dfrac{1}{M} \sum_{j=1}^{M} \norm{ \dfrac{\mathrm{d}NN_{\theta}(t, \textbf{E})}{\mathrm{d}E_{j}}}_2^2 \\[2ex]
\end{aligned}
\label{eqn:reg}
\end{equation}
where M is the total number of environmental parameters, $E_{j}$ is one of the environmental parameters considered. Consequently, the total loss function is $\mathcal{L}_{reg} = \mathcal{L}_{NO} + \lambda \mathcal{L}_{smooth}$
where $\lambda$ is the weighting factor for the regularisation term. This model will be denoted as regularised LiFe-net.

\subsection{Neural Operator with Time Stability Loss}
\begin{figure}[htbp]
\centerline{\includegraphics[width=1\textwidth]{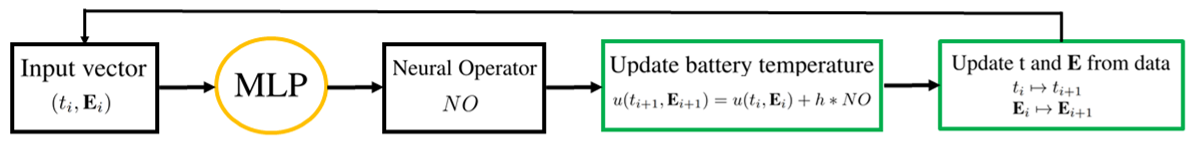}}
\caption{Training pipeline of LiFe-net (time stability). See Algorithm \ref{alg:training} and \ref{alg:inference}.}
\label{fig:training_pipeline}
\end{figure}

In this subsection we incorporate the temperature evolution into the training objective in terms of a time-stability loss $\mathcal{L}_{TS}$, by introducing the Euler forward method into the training loop. 
It allows us to recover the complete time evolution of the temperature for each drive during training and do a multi-time-step optimization between the predicted and the ground-truth temperatures at once (see Equation \ref{eqn:nostb}). This enforces local smoothness like $\mathcal{L}_{reg}$ as well as consistency with our training data.
\begin{equation}
\begin{aligned}
	\mathcal{L}_{TS} &= \sum_{i=1}^{N-1} \norm{ u_{gt}^i - (NN_{\theta}(t_{i-1}, \textbf{E}_{i-1})*h+u_{pred}^{i-1}) }_2^2
\end{aligned}
\label{eqn:nostb}
\end{equation}
where $u_{gt}^{i}$  and $u_{pred}^{i}$ are the ground-truth and predicted temperatures at time step $i$, respectively. N is the number of time steps for the particular drive session. This model will be denoted as LiFe-net (time-stability) in this article. The training pipeline of this model is illustrated in Fig.\ref{fig:training_pipeline}

\subsection{Surrogate Model of Charging Statistics}
The correlation of battery temperature to charging speed $c_p$ and charging time $t_{charge}$ is modeled by a linear regression model, viz.
\begin{equation} \label{eq:surrogate}
\begin{split}
    c_p &= c_{soc} C_{bat} + c_{T} u_i(t_i, E_i) + o \\
    t_{charge} &= c_{soc}^s C_{bat}^s + c_{soc}^e C_{bat}^e + c_{T}^{t} u_i(t_i, E_i) + o^{t} \\
\end{split}
\end{equation}

with $c_{soc}$ explaining the contribution of state-of-charge $C_{bat}$ and $c_{T}$ scaling the contribution of battery temperature $u_i(t_i, E_i)$. Charging time is also modeled by coefficients $c_{soc}^s, c_{soc}^e$ corresponding to state-of-charge at the beginning $C_{bat}^s$ and end $C_{bat}^e$ of the charging session. $c_{T}^{t}$ models the contribution of battery temperature while $o, o^{t}$ introduce offsets to the predicted charging time and power. The connection of this surrogate model with our neural operator now allows for advanced driver's assistance in deciding whether to heat the battery or not (see Fig. \ref{fig:pwr} in Appendix A).

\section{Results and Discussion}
The full dataset consists of 445 000 data measurements collected during 99 drives under different environmental conditions. The data was split with respect to individual drives leaving 92 drives for training and 7 for testing. Hence, the split for the training and test dataset is  80\% and 20\%, respectively. Table \ref{table:Data} summarizes the range for parameters used. All models were trained on the Taurus HPC system at TU Dresden using Tesla V100-SXM2-32GB GPU. Furthermore, we validate the performance of the models on the test data using MSE, MAE and relative error between predicted and ground-truth temperature values. The corresponding dataset and the implementation code are available at the following GitHub repository: \url{https://github.com/Jeyhun1/LiFe-net.git}

\subsection{Prediction accuracy}
\begin{figure}
\centering
\begin{subfigure}{0.4\textwidth}
    \includegraphics[width=\textwidth]{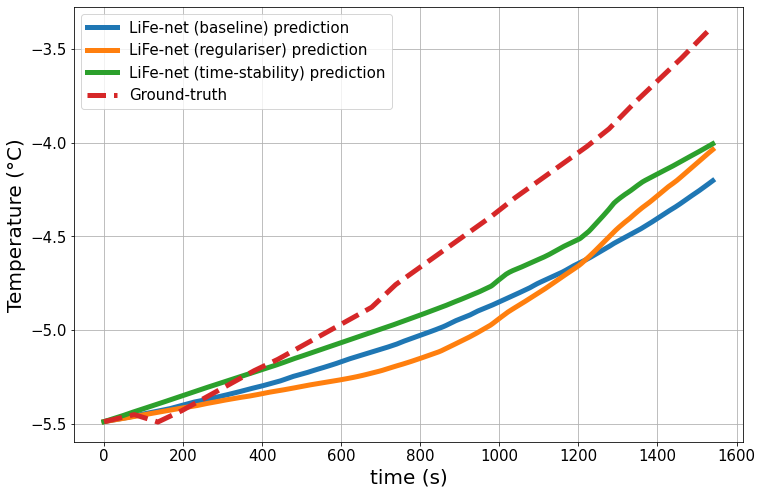}
    \caption{}
    \label{fig:first}
\end{subfigure}
\hfill
\begin{subfigure}{0.4\textwidth}
    \includegraphics[width=\textwidth]{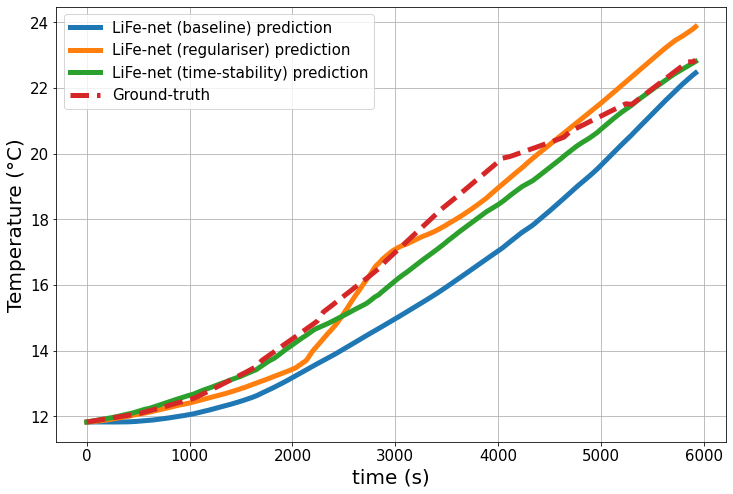}
    \caption{}
    \label{fig:second}
\end{subfigure}
        
\caption{(a) Time-stability loss is improving the accuracy at the interface of our training data range (here: winter). (b) All models perform well on test data that is well represented by our training data.}
\label{fig:tstb}
\end{figure}

We performed hyperparameter optimization of all discussed models using Weights and Biases library \cite{wandb}. The results summarised in Table \ref{table:lambda} show that regularised LiFe-net with $\lambda$ parameter value of 0.1 outperforms the baseline model where $\lambda$ = 0. 
Furthermore, Tables \ref{table:lifenetreg} and \ref{table:lifenettstb} (See Appendix A) display MSE  for the different number of hidden layers and different numbers of neurons per layer for both regularised LiFe-net and time-stability LiFe-net, respectively.
Finally, Table \ref{table:testdata} summarises the prediction accuracy of best performing models for each method considered in this work averaged over seven test data. Furthermore, comparison plots of all three models on individual test datasets are shown in Fig. \ref{fig:tstb}. The results show that the LiFe-net model trained with the time stability loss function performs the best, especially on data at the interface of the training data range (see Figure \ref{fig:first}). This can be explained by the multi-time step optimisation during training which the other two methods lack. However, this comes at the cost of considerably slower training time (see Table \ref{table:testdata}).

\subsection{Learning Charging Statistics from Temperature Data}
The coefficients of our surrogate model explaining charging statistics as of Equation \ref{eq:surrogate} have been estimated by solving the corresponding least-squares objective function by normal equations. The regression model of $c_p$ got a $R^2 = 0.726$ while $t_{charge}$ was explained even better since $R^2 = 0.884$. The coefficients can be found in Table \ref{tbl:surrogateModel}. It can be observed that the state-of-charge at start of charging session is important for predicting both peak charging power ($c_{soc}=-0.9557$, $p=0.015$) as well as -time ($c_{soc}^s = -0.6213$, $p=0.018$). The reciprocal relationship implies that lower state-of-charge results in higher maximum charging power and therefore fast charging times since electrons can be transferred easier into the battery. Electrochemical processes also react faster in warmer environmental conditions which are reflected by the significant contribution of $c_{T}$ and $c_{T}^{\Delta_c}$.

\begin{table}[htbp!]
\caption{Prediction accuracy of LiFe-net averaged on 7 test data.}
\centering
\label{table:testdata}
\begin{tabular}{l c c c c c}
\toprule
\multirow{1}{*}{Model} & 
\multicolumn{1}{c}{MAE} & 
\multicolumn{1}{c}{MSE} & 
\multicolumn{1}{c}{Relative error (\%)} &
\multicolumn{1}{c}{time per epoch (s)} \\ 

\midrule
LiFe-net (baseline) & 0.7068 & 0.9342  & 4.6452 & \textbf{104.89}  \\

LiFe-net ($\mathcal{L}_{reg}$) & 0.5445 & 0.6312 & 3.6223 & 123.46 \\

LiFe-net ($\mathcal{L}_{TS}$) & \textbf{0.3845} & \textbf{0.2692} &  \textbf{2.7715} & 1151.25\\


\bottomrule
\end{tabular}
\end{table}

\section{Conclusion and broader impact}

In this work, we introduced three variations of a Neural Operator that is learning the temperature evolution of a temperature probe attached to the surface of an EV battery depending on environmental conditions. We demonstrated that the LiFe-net trained with time-stability loss where we optimized the solution operator in a multi-timestep fashion is the most accurate. This information is then passed into a surrogate model that relates the measured temperature to charging statistics such as charging time and maximum charging power. Hence, this model enables EV manufacturers to get real-time never-before accessible data regarding battery sensitivities to numerous environmental conditions, which will give insights into designing more efficient battery systems in the future.

\bibliography{references}

\section*{Checklist}


\begin{enumerate}

\item For all authors...
\begin{enumerate}
  \item Do the main claims made in the abstract and introduction accurately reflect the paper's contributions and scope?
    \answerYes{}
  \item Did you describe the limitations of your work?
    \answerNA{}
  \item Did you discuss any potential negative societal impacts of your work?
    \answerNA{}
  \item Have you read the ethics review guidelines and ensured that your paper conforms to them?
    \answerYes{}
\end{enumerate}

\item If you are including theoretical results...
\begin{enumerate}
  \item Did you state the full set of assumptions of all theoretical results?
    \answerNA{}
        \item Did you include complete proofs of all theoretical results?
    \answerNA{}
\end{enumerate}

\item If you ran experiments...
\begin{enumerate}
  \item Did you include the code, data, and instructions needed to reproduce the main experimental results (either in the supplemental material or as a URL)?
    \answerYes{Section 3}
  \item Did you specify all the training details (e.g., data splits, hyperparameters, how they were chosen)?
    \answerYes{See section 3}
        \item Did you report error bars (e.g., with respect to the random seed after running experiments multiple times)?
    \answerNA{}
        \item Did you include the total amount of compute and the type of resources used (e.g., type of GPUs, internal cluster, or cloud provider)?
    \answerYes{See section 3}
\end{enumerate}

\item If you are using existing assets (e.g., code, data, models) or curating/releasing new assets...
\begin{enumerate}
  \item If your work uses existing assets, did you cite the creators?
    \answerNA{}
  \item Did you mention the license of the assets?
   \answerNA{}
  \item Did you include any new assets either in the supplemental material or as a URL?
    \answerNA{}
  \item Did you discuss whether and how consent was obtained from people whose data you're using/curating?
    \answerNA{}
  \item Did you discuss whether the data you are using/curating contains personally identifiable information or offensive content?
    \answerNA{}
\end{enumerate}

\item If you used crowdsourcing or conducted research with human subjects...
\begin{enumerate}
  \item Did you include the full text of instructions given to participants and screenshots, if applicable?
    \answerNA{}
  \item Did you describe any potential participant risks, with links to Institutional Review Board (IRB) approvals, if applicable?
    \answerNA{}
  \item Did you include the estimated hourly wage paid to participants and the total amount spent on participant compensation?
    \answerNA{}
\end{enumerate}

\end{enumerate}

\appendix

\section{Appendix}

\begin{figure}[htbp!]
\centering
\includegraphics[width=.45\textwidth]{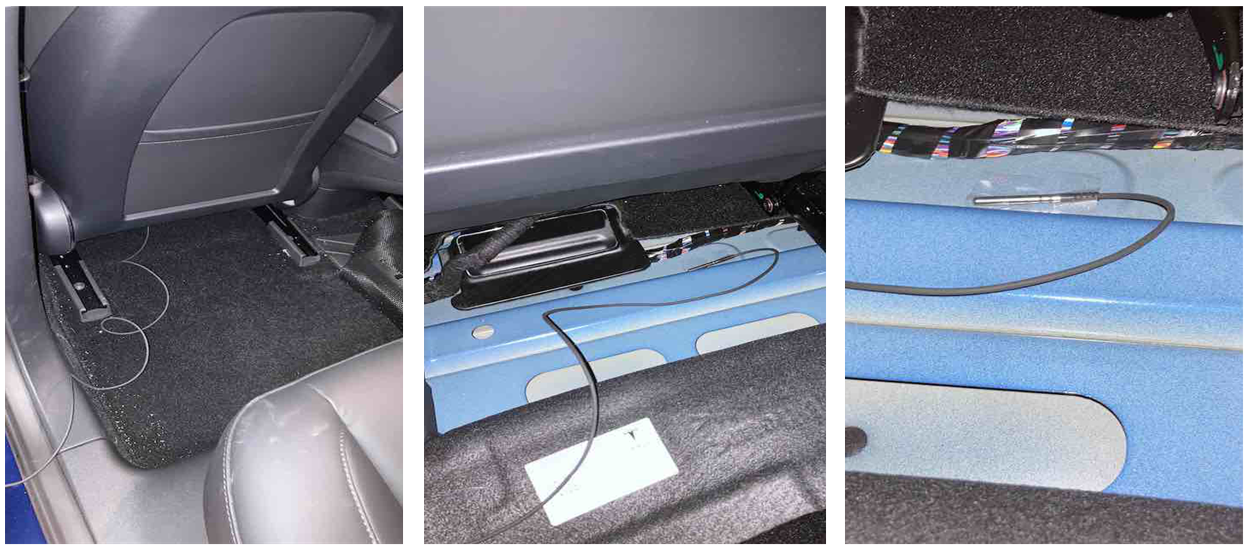} 
\caption{Installation of the sensor on the floor of the cabin.
}
\label{fig:sensorinst}
\end{figure}

\begin{figure}[htbp!] 
\centerline{\includegraphics[width=.35\textwidth]{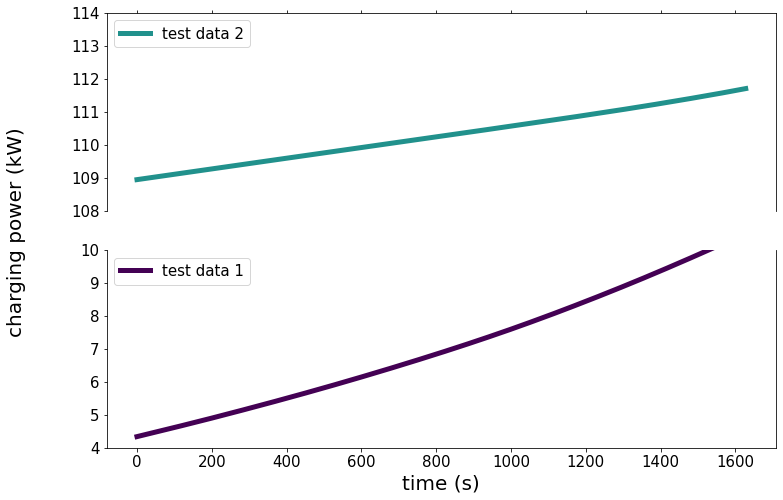}}
\caption{The connection of LiFe-net with our surrogate model now allows us to monitor and predict evolution of peak charging power during drives. Peak charging power is strongly reduced in winter as can be seen in test data 1 (purple).}
\label{fig:pwr}
\end{figure}

\begin{table}[htbp!]
\caption{The value range of parameters in training and test dataset.}
\centering
\label{table:Data}
\begin{tabular}{l c c }
\toprule
\multirow{1}{*}{Parameter} & \multicolumn{1}{c}{Training data} & \multicolumn{1}{c}{Test data} \\ 

\midrule
power (kW) & [-68, 166] & [-67, 152]   \\

speed (km/h) & [0, 167] & [0, 166]   \\

battery level (\%) & [10, 100] & [22, 95]   \\

outside temperature ($^\circ$C) & [-9.5, 35.5] & [-10, 29.75] \\

battery temperature ($^\circ$C) & [-5.79, 33.9] & [-5.49, 33.85]
\\
\bottomrule
\end{tabular}
\end{table}

\begin{table}[htbp!]
\caption{MSE for regularised LiFe-net with different $\lambda$ values}
\centering
\label{table:lambda}
\begin{tabular}{l c c c c c c c c c c c}
\toprule
\multirow{1}{*}{$\lambda$} &\multicolumn{1}{c}{0} &\multicolumn{1}{c}{1e-6} &\multicolumn{1}{c}{1e-4} &\multicolumn{1}{c}{0.01} &\multicolumn{1}{c}{0.1} &\multicolumn{1}{c}{1} &\multicolumn{1}{c}{10} &\multicolumn{1}{c}{100}  &\multicolumn{1}{c}{100}  &\multicolumn{1}{c}{10000} \\ 
\midrule
MSE & 0.9342 & 0.934 & 0.9313 & 0.9268 & \textbf{0.8349}  & 0.9317 & 0.9076 & 1.233 & 3.165 & 4.56
\\

\bottomrule
\end{tabular}
\end{table}

\begin{table}[htbp!]
\caption{MSE for different regularised LiFe-net architectures with $\lambda$=0.1 }
\centering
\label{table:lifenetreg}
\begin{tabular}{l c c c c }
\toprule
\multirow{2}{*}{Hidden size} & \multicolumn{4}{c}{Number of hidden layers} \\
\cmidrule{2-5}
  & 2 & 4 & 6  & 8  \\ 
\midrule
10 & 0.7578 & 0.7787 &  0.7929 &  1.267  \\
20 & 0.6981 & 1.132  & 0.8298 & 0.9033 \\
50 & 0.7632 & 0.827 & 0.636 & \textbf{0.6312} \\
100 & 0.6325 & 0.8349 & 0.7817 & 0.8284\\

\bottomrule
\end{tabular}
\end{table}

\begin{table}[htbp!]
\caption{MSE for different LiFe-net (time-stability) architectures }
\centering
\label{table:lifenettstb}
\begin{tabular}{l c c c c }
\toprule
\multirow{2}{*}{Hidden size} & \multicolumn{4}{c}{Number of hidden layers} \\
\cmidrule{2-5}
  & 2 & 4 & 6  & 8  \\ 
\midrule
10 & 0.4734 & 1.631 &  1.51 &  0.8112  \\
20 & 0.7086 & 0.6756  & 0.5135 & 0.3114 \\
50 & 0.5767 & 0.5748 & 0.5943 & 0.4667 \\
100 & 0.6005 & 0.5296 & 0.4614 & \textbf{0.2692}\\

\bottomrule
\end{tabular}
\end{table}

\begin{table}[htbp!]
\caption{Coefficients of Surrogate Model with $^{\ast}$ denoting significant coefficients at $p<0.5$.}
\label{tbl:surrogateModel}
\centering
\begin{tabular}{c c c c}
$c_{soc}$ & $c_{T}$ & $o$  \\ 
 \toprule
$-0.6109^{\ast}$ & $3.7923^{\ast}$ & $39.7772^{\ast}$ \\
\bottomrule
\\
$c_{soc}^s$ & $c_{soc}^e$ & $c_{T}^{t}$ & $o^{t}$ \\
\toprule
 $-0.4505^{\ast}$ & $0.7690^{\ast}$ & $-0.6267^{\ast}$ & $-6.9412$ \\
\bottomrule
\end{tabular}
\end{table}

\begin{algorithm}[htbp!]
	\caption{Training of LiFe-net with time stability loss}
	\begin{algorithmic}[1]
		\For {$epoch$ in $0, 1,\ldots,N$}
			\For {M number of batches} 
			    \State Set initial condition for env.parameters in 'input'
            \For{$j$ in  $0, 1,\ldots,T.shape-1$}
                \State NO = mlp(input[j])
                \State u\_pred[j+1]=u\_pred[j]+NO * step\_size[j]
                \State Update t, u\_pred and env. parameters in the 'input'
            \EndFor
            \State loss = MSE(u\_pred, u\_gt)
            \State loss.backward()
            \State optimizer.step()
			\EndFor
		\EndFor
	\end{algorithmic} 
\label{alg:training}
\end{algorithm}
\vspace{-100mm}
\begin{algorithm}[htbp!]
	\caption{Inference phase of LiFe-net} 
	\begin{algorithmic}[1]
	\State Set initial condition for env.parameters in 'input'
		\For {$i$ in $0,1,\ldots,T.shape-1$}
        \State NO = model(input[i])
        \State u\_pred[i+1]=u\_pred[i]+NO * step\_size[i]
        \State Update t, u\_pred and env. parameters in the 'input'
		\EndFor
	\end{algorithmic} 
\label{alg:inference}
\end{algorithm}

\end{document}